%% file: main.tex
\def\year{2020}\relax
\documentclass[letterpaper]{article} 
\usepackage{aaai20}  
\usepackage{times}  
\usepackage{helvet} 
\usepackage{courier}  
\usepackage[hyphens]{url}  
\usepackage{graphicx} 
\usepackage{soul}
\usepackage{amsmath}
\usepackage{booktabs}
\usepackage{algorithm}
\usepackage{algorithmic}
\usepackage{makecell}
\usepackage{color}
\usepackage{graphicx}  
\title{Uncertainty-Aware Search Framework for Multi-Objective Bayesian Optimization}
\author{Syrine Belakaria, Aryan Deshwal, Nitthilan Kannappan Jayakodi, Janardhan Rao Doppa\\ 
School of EECS, Washington State University\\
\text{\{syrine.belakaria, aryan.deshwal, n.kannappanjayakodi, jana.doppa\}@wsu.edu}
}
 
\begin{document}

\maketitle

\begin{abstract}
We consider the problem of multi-objective (MO) blackbox
optimization using expensive function evaluations, where the
goal is to approximate the true Pareto set of solutions while
minimizing the number of function evaluations. For example,
in hardware design optimization, we need to find the designs
that trade-off performance, energy, and area overhead using
expensive simulations. We propose a novel uncertainty-aware
search framework referred to as USeMO to efficiently select the
sequence of inputs for evaluation to solve this problem. The
selection method of USeMO consists of solving a cheap MO
optimization problem via surrogate models of the true functions
to identify the most promising candidates and picking the
best candidate based on a measure of uncertainty. We also
provide theoretical analysis to characterize the efficacy of
our approach. Our experiments on several synthetic and six
diverse real-world benchmark problems show that USeMO
consistently outperforms the state-of-the-art algorithms.
\end{abstract}

\section{Introduction}

Many engineering and scientific applications involve making design choices to optimize multiple objectives. Some examples include tuning the knobs of a compiler to optimize performance and efficiency of a set of software programs; and designing new materials to optimize strength, elasticity, and durability. There are two 
challenges in solving these kind of optimization problems: {\bf 1)} The objective functions are unknown and we need to perform expensive experiments to evaluate each candidate design. For example, performing computational simulations and physical lab experiments for compiler optimization and material design applications respectively. {\bf 2)} The objectives are conflicting in nature and all of them cannot be optimized simultaneously. Therefore, we need to find the {\em Pareto optimal} set of solutions. A solution is called Pareto optimal if it cannot be improved in any of the objectives without compromising some other objective. The overall 
goal is to approximate the true Pareto set while minimizing the number of function evaluations. 

Bayesian Optimization (BO) \cite{shahriari2016taking} is an effective framework to solve blackbox optimization problems with expensive function evaluations. The key idea behind BO is to
build a cheap surrogate model (e.g., Gaussian Process \cite{williams2006gaussian}) using the real experimental evaluations; and employ it to intelligently select the sequence of function evaluations using an acquisition function, e.g., expected improvement (EI). There is a large body of literature on single-objective BO algorithms \cite{shahriari2016taking} and their applications including hyper-parameter tuning of machine learning methods \cite{snoek2012practical,kotthoff2017auto}. However, there is relatively less work on the more challenging problem of BO for multiple objectives. 

Prior work on multi-objective BO is lacking in the following ways. Many algorithms reduce the problem to single-objective optimization by designing appropriate acquisition functions, e.g., expected improvement in Pareto hypervolume \cite{knowles2006parego,emmerich2008computation}. Unfortunately, this choice is sub-optimal as it is hard to capture the trade-off between multiple objectives and can potentially lead to aggressive exploitation behavior. Additionally, algorithms to optimize Pareto Hypervolume (PHV) based acquisition functions scale poorly as the number of objectives and dimensionality of input space grows. PESMO is a state-of-the-art information-theoretic approach that relies on the principle of input space entropy search \cite{hernandez2016predictive}. However, it is computationally expensive to optimize the acquisition function behind PESMO. A series of approximations are performed to improve the efficiency potentially at the expense of accuracy.

In this paper, we propose a novel {\bf U}ncertainty-aware {\bf Se}arch framework for optimizing {\bf M}ultiple {\bf O}bjectives (USeMO) to overcome the drawbacks of prior methods. The key insight behind USeMO is a two-stage search procedure to improve the accuracy and computational-efficiency of sequential decision-making under uncertainty for selecting candidate inputs for evaluation. USeMO selects the inputs for evaluation as follows. First, it solves a cheap MO optimization problem defined in terms of the acquisition functions (one for each unknown objective) to identify a list of promising candidates. Second, it selects the best candidate from this list based on a measure of uncertainty. Unlike prior methods, USeMO has several advantages: a) Does not reduce to single objective optimization problem; b) Allows to leverage a variety of acquisition functions designed for single objective BO; c) Computationally-efficient to solve MO problems with many objectives; and d) Improved uncertainty management via two-stage search procedure to select the candidate inputs for evaluation. 

\vspace{0.8ex}

\noindent {\bf Contributions.} The main contributions of this paper are:
\begin{itemize}
    
    \item Developing a principled search-based BO framework referred as USeMO to solve multi-objective blackbox optimization problems.
    
    \item Theoretical analysis of the USeMO framework in terms of asymptotic regret bounds.
    
    \item Comprehensive experiments over synthetic and {\em six} diverse real-world benchmark problems to show the accuracy and efficiency improvements over existing methods. 
\end{itemize}

\section{Background and Problem Setup}

\vspace{0.5ex}

\noindent {\bf Bayesian Optimization Framework.} 
Let $\mathcal{X} \subseteq \Re^d$ be an input space. We assume an unknown real-valued objective function $F: \mathcal{X} \mapsto \Re$, which can evaluate each input $x \in \mathcal{X}$ to produce an evaluation $y$ = $F(x)$.  Each evaluation $F(x)$ is expensive in terms of the consumed resources. The main goal is to find an input $x^* \in \mathcal{X}$ that approximately optimizes $F$ via a limited number of function evaluations. BO algorithms learn a cheap surrogate model from training data obtained from past function evaluations. They intelligently select the next input for evaluation by trading-off exploration and exploitation to quickly direct the search towards optimal inputs. The three key elements of BO framework are:

 {\bf 1) Statistical Model} of 
 $F(x)$. {\em Gaussian Process (GP)} \cite{williams2006gaussian} is the most commonly used model. A GP over a space $\mathcal{X}$ is a random process from $\mathcal{X}$ to $\Re$. It is characterized by a mean function $\mu : \mathcal{X} \times \mathcal{X} \mapsto \Re$ and a covariance or kernel function $\kappa$. If a function $F$ is sampled from GP($\mu$, $\kappa$), then $F(x)$ is distributed normally $\mathcal{N}(\mu(x), \kappa(x,x))$ for a finite set of inputs from $x \in \mathcal{X}$.
 
 {\bf 2) Acquisition Function} (\textsc{Af}) to score the utility of evaluating a candidate input $x \in \mathcal{X}$ based on the statistical model. Some popular acquisition functions include expected improvement (EI), upper confidence bound (UCB), lower confidence bound (LCB), and Thompson sampling (TS). For the sake of completeness, we formally define the acquisition functions employed in this work noting that any other acquisition function can be employed within USeMO. 
 \begin{align}
    & UCB(x)=\mu(x)+\beta^{1/2} \sigma(x)\\\label{ucbeq}
    & LCB(x)=\mu(x)-\beta^{1/2} \sigma(x)\\\label{lcbeq}
    & TS(x)=f(x) ~ with ~ f(.) \sim GP\\
    & EI(x)= \sigma(x)(\alpha\Phi(\alpha)+\phi(\alpha)) , ~ \alpha=\frac{\tau-\mu(x)}{\sigma(x)}
\end{align}
where $\mu(x)$ and $\sigma(x)$ correspond to the mean and standard deviation of the prediction from statistical model, and represent exploitation and exploration scores respectively; $\beta$ is a parameter that balances exploration and exploitation; $GP$ is the statistical model learned from past observations; $\tau$ is the best uncovered input; and $\Phi$ and $\phi$ are the CDF and PDF of normal distribution respectively.

{\bf 3) Optimization Procedure} to select the best scoring candidate input according to \textsc{Af} via statistical model, e.g., DIRECT \cite{jones1993lipschitzian}. 

\vspace{1.5ex}

\noindent {\bf Multi-Objective Optimization (MOO) Problem.}  Without loss of generality, our goal is to minimize $k \geq 2$ real-valued objective functions $F_1(x), F_2(x),\cdots,F_k(x)$ over continuous space  $X \subseteq \Re^d$. Each evaluation of an input $x \in \mathcal{X}$ produces a vector of objective values $Y$ = $(y_1, y_2,\cdots,y_k)$ where $y_i$ = $F_i(x)$ for all $i \in \{1,2, \cdots, k\}$. We say that a point $x$ {\em Pareto-dominates} another point $x'$ if $F_i(x) \leq F_i(x') \hspace{1mm} \forall{i}$ and there exists some $j \in \{1, 2, \cdots,k\}$ such that $F_j(x) < F_j(x')$. The optimal solution of MOO problem is a set of points $\mathcal{X}^* \subset \mathcal{X}$ such that no point $x' \in \mathcal{X} \setminus \mathcal{X}^*$ Pareto-dominates a point $x \in \mathcal{X}^*$. The solution set $\mathcal{X}^*$ is called the Pareto set and the corresponding set of function values is called the Pareto front. Our goal is to approximate $\mathcal{X}^*$ while minimizing the number of function evaluations.

\section{Related work}

There is a family of model-based MO optimization algorithms that reduce the problem to single-objective optimization. ParEGO method \cite{knowles2006parego} employs random scalarization for this purpose: scalar weights of $k$ objective functions are sampled from a uniform distribution to construct a single-objective function and expected improvement is employed as the acquisition function to select the next input for evaluation. ParEGO is simple and fast, but more advanced approaches often outperform it. Recently, \cite{paria2018flexible} proposed a scalarization based method focusing on a specialized setting, where preference over objective functions is specified as input. The preference is expressed in terms of the values of the scalars.
Many methods optimize the Pareto hypervolume (PHV) metric \cite{emmerich2008computation} that captures the quality of a candidate Pareto set. This is done by extending the standard acquisition functions to PHV objective, e.g., expected improvement in PHV \cite{emmerich2008computation} and probability of improvement in PHV \cite{picheny2015multiobjective}. Unfortunately, algorithms to optimize PHV based acquisition functions scale very poorly and are not feasible for more than two objectives. To improve scalability, methods to reduce the search space are also explored \cite{ponweiser2008multiobjective}. A common drawback of this family of algorithms is that reduction to single-objective optimization can be sub-optimal: it is hard to capture the trade-off between multiple objectives and can potentially lead to more exploitation behavior.

PAL \cite{zuluaga2013active} and PESMO \cite{hernandez2016predictive} are principled algorithms based on information theory. PAL tries to classify the input points based on the learned models into three categories: Pareto optimal, non-Pareto optimal, and uncertain. In each iteration, it selects the candidate input for evaluation towards the goal of minimizing the size of the uncertain set. PAL provides theoretical guarantees, but it is only applicable for input space $\mathcal{X}$ with finite set of discrete points. PESMO is a state-of-the-art method based on entropy optimization. It iteratively selects the input that maximizes the information gained about the true Pareto set. Unfortunately, it is computationally expensive to optimize the acquisition function employed in PESMO. Some approximations are performed to improve the efficiency of acquisition function optimization, but can potentially degrade accuracy and result in loss of information-theoretical advantage. MESMO \cite{MESMO} is a concurrent work based on output space entropy search that improves over PESMO.

In the domain of analog circuit design optimization, \cite{lyu2018multi} developed a technique that conducts an optimization over the posterior means of the GPs using LCB acquisition function. It is an application-specific solution, whereas we show that USeMO generalizes for six diverse application domains including hyper-parameter tuning in neural networks, compiler settings, network-on-chip, and materials design. Additionally, we show consistently better performance using multiple acquisition functions. 

\section{Uncertainty-Aware Search Framework}
\label{section4}

In this section, we provide the details of USeMO framework for solving multi-objective optimization problems. First, we provide an overview of USeMO followed by the details of its two main components. Subsequently, we provide theoretical analysis of USeMO in terms of asymptotic regret bounds.


\subsection{Overview of USeMO Framework} 

\begin{figure}[h]
    \centering
    \includegraphics[scale=0.4]{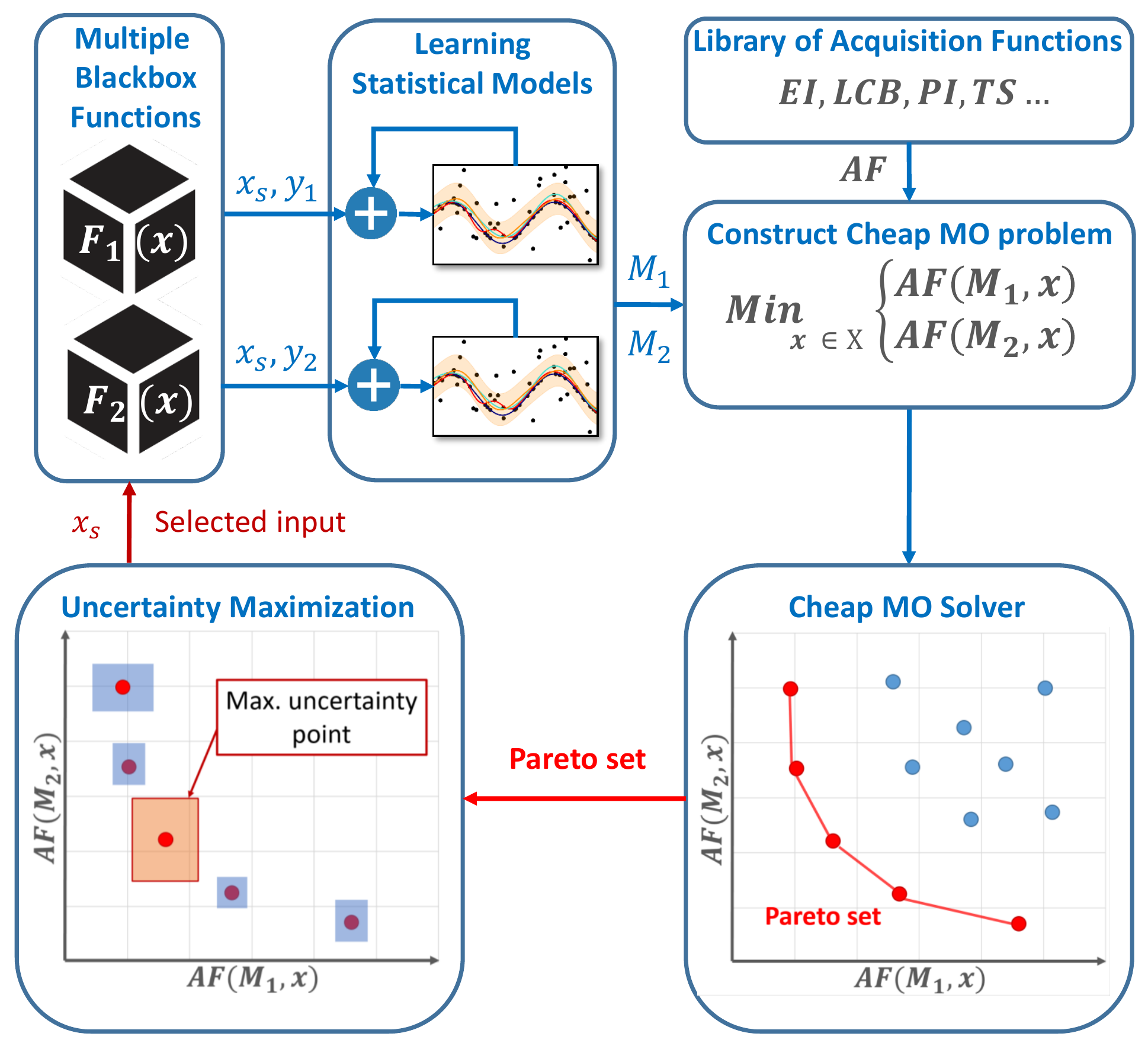}
    \caption{Overview of the USeMO framework for two objective functions ($k$=2). We build statistical models $\mathcal{M}_1$, $\mathcal{M}_2$ for the two objective functions $F_1(x)$ and $F_2(x)$. In each iteration, we perform the following steps. First, we construct a cheap MO problem using the statistical models $\mathcal{M}_1$ and $\mathcal{M}_2$ and an input acquisition function \textsc{Af}: $\min_{x \in \mathcal{X}} \; (\textsc{Af}(\mathcal{M}_1,x), \textsc{Af}(\mathcal{M}_2,x))$ and employ a cheap MO solver to find the promising candidate inputs in the form of Pareto set. Second, we select the best candidate input $x_s$ from the Pareto set based on a measure of uncertainty. Finally, we evaluate the functions for $x_s$ to get $Y_s$=$(y_1, y_2)$ and update the statistical models using the new training example.}
    \label{fig:USeMO}
\end{figure}

As shown in Figure~\ref{fig:USeMO}, USeMO is a iterative
algorithm that involves four key steps. First, We build statistical models $\mathcal{M}_1, \mathcal{M}_2,\cdots,\mathcal{M}_k$ for each of the $k$ objective functions from the training data in the form of past function evaluations. Second, we select a set of promising candidate inputs $\mathcal{X}_p$ by solving a cheap MO optimization problem defined using the statistical models. Specifically, multiple objectives of the cheap MO problem correspond to $\textsc{Af}(\mathcal{M}_1,x), \textsc{Af}(\mathcal{M}_2,x),\cdots, \textsc{Af}(\mathcal{M}_k,x)$ respectively.
Any standard acquisition function \textsc{Af} from single-objective BO (e.g., EI, TS) can be used for this purpose. The Pareto set $\mathcal{X}_p$ corresponds to the inputs with different trade-offs in the utility space for $k$ unknown functions. Third, we select the best candidate input $x_s \in \mathcal{X}_p$  from the Pareto set that maximizes some form of uncertainty measure for evaluation. Fourth, the selected input $x_s$ is used for evaluation to get the corresponding function evaluations: $y_1$=$F_1(x_s)$, $y_2$=$F_2(x_s)$,$\cdots$,$y_k$=$F_k(x_s)$. The next iteration starts after the statistical models $\mathcal{M}_1, \mathcal{M}_2,\cdots,\mathcal{M}_k$ are updated using the new training example: input is $x_s$ and output is ($y_1,y_2,\cdots,y_k$). Algorithm~\ref{alg:USeMO} provides the algorithmic pseudocode for USeMO.

\vspace{0.8ex}

\noindent {\bf Advantages.} USeMO has many advantages over prior methods. {\bf 1)} Provides flexibility to plug-in any acquisition function for single-objective BO. This allows us to leverage existing acquisition functions including EI, TS, and LCB. {\bf 2)} Unlike methods that reduce to single-objective optimization, USeMO has a better mechanism to handle uncertainty via a two-stage procedure to select the next candidate for evaluation: pareto set obtained by solving cheap MO problem contains all promising candidates with varying trade-offs in the utility space and the candidate with maximum uncertainty from this list is selected. {\bf 3)} Computationally-efficient to solve MO problems with many objectives. 

\begin{algorithm}[h]
\caption{USeMO Framework}
\label{alg:USeMO}
\textbf{Input}: $\mathcal{X}$, input space; $F_1(x), F_2(x),\cdots,F_k(x)$, $k$ blackbox objective functions; \textsc{Af}, acquisition function; and $T_{max}$, maximum no. of iterations
\begin{algorithmic}[1] 
\STATE Initialize training data of function evaluations $\mathcal{D}$
\STATE Initialize statistical models $\mathcal{M}_1, \mathcal{M}_2,\cdots, \mathcal{M}_k$ from $\mathcal{D}$
\FOR{each iteration $t=1$ to $T_{max}$}
\STATE // Solve cheap MO problem  with objectives $\textsc{Af}(\mathcal{M}_1,x),\cdots, \textsc{Af}(\mathcal{M}_k,x)$ to get candidate inputs
\STATE $\mathcal{X}_{p} \leftarrow \min_{x \in \mathcal{X}} \; (\textsc{Af}(\mathcal{M}_1,x),\cdots, \textsc{Af}(\mathcal{M}_k,x))$ 
\STATE // Pick the candidate input with maximum uncertainty
\STATE Select $x_{t+1} \leftarrow \arg max_{x\in \mathcal{X}_{p}} \; U_{\beta_t}(x)$
\STATE Evaluate $x_{t+1}: Y_{t+1} \leftarrow (F_1(x_{t+1}),\cdots,F_k(x_{t+1}))$
\STATE Aggregate data: $\mathcal{D} \leftarrow \mathcal{D} \cup \{(x_{t+1}, Y_{t+1})\}$ 
\STATE Update models $\mathcal{M}_1, \mathcal{M}_2,\cdots, \mathcal{M}_k$ using $\mathcal{D}$ 
\STATE $t \leftarrow t+1$
\ENDFOR
\STATE \textbf{return} Pareto set and Pareto front of $\mathcal{D}$
\end{algorithmic}
\end{algorithm}

\subsection{Key Algorithmic Components of USeMO} 

The two main algorithmic components of USeMO framework are: selecting most promising candidate inputs by solving a cheap MO problem and picking the best candidate via uncertainty maximization. We describe their details below.

\vspace{0.8ex}

\noindent {\bf Selection of promising candidate inputs.} We employ the statistical models $\mathcal{M}_1, \mathcal{M}_2,\cdots, \mathcal{M}_k$ towards the goal of selecting promising candidate inputs as follows. Given a acquisition function \textsc{Af} (e.g., EI), we construct a cheap multi-objective optimization problem with objectives $\textsc{Af}(\mathcal{M}_1,x), \textsc{Af}(\mathcal{M}_2,x),\cdots, \textsc{Af}(\mathcal{M}_k,x)$, where $\mathcal{M}_i$ is the statistical model for unknown function $F_i$. Since we present the framework as minimization for the sake of technical exposition, all AFs will be minimized. The Pareto set $\mathcal{X}_p$ obtained by solving this cheap MO problem represents the most promising candidate inputs for evaluation.
\begin{align}
\label{eq2}
\mathcal{X}_p\leftarrow \min_{x \in \mathcal{X}} \; (\textsc{Af}(\mathcal{M}_1,x),\cdots, \textsc{Af}(\mathcal{M}_k,x))
\end{align}
Each acquisition function \textsc{Af}($\mathcal{M}_i,x$) is dependent on the corresponding surrogate model $\mathcal{M}_i$ of the unknown objective function $F_i$. Hence, each acquisition function will carry the information of its associated objective function. As iterations progress, using more training data, the models $\mathcal{M}_1, \mathcal{M}_2,\cdots, \mathcal{M}_k$ will better mimic the true objective functions $F_1, F_2,\cdots,F_k$. Therefore, the Pareto set of the acquisition function space (solution of Equation~\ref{eq2}) becomes closer to the Pareto set of the true functions $\mathcal{X}^*$ with increasing iterations.
Intuitively, the acquisition function \textsc{Af}($\mathcal{M}_i,x$) corresponding to unknown objective function $F_i$ tells us the utility of a point $x$ for optimizing $F_i$. The input minimizing \textsc{Af}($\mathcal{M}_i,x$) has the highest utility for $F_i$, but may have a lower utility for a different function $F_j$ ($j \neq i$). The utility of inputs for evaluation of $F_j$ is captured by its own acquisition function \textsc{Af}($\mathcal{M}_j,x$). Therefore, there is a trade-off in the utility space for all $k$ different functions. The Pareto set $\mathcal{X}_p$ obtained by simultaneously optimizing acquisition functions for all $k$ unknown functions will capture this utility trade-off. As a result, each input $x \in \mathcal{X}_p$ is a promising candidate for evaluation towards the goal of solving MOO problem. 
USeMO employs the same acquisition function for all $k$ objectives. The main reason is to give equivalent evaluation for all functions in the Pareto front (PF) at each iteration. If we use different AFs for different objectives, the sampling procedure would be different. Additionally, the values of various AFs can have considerably different ranges. Thus, this can result in an unbalanced trade-off between functions in the cheap PF leading to the same unbalance in our final PF.
\vspace{0.5ex}

{\bf Cheap MO solver.} We employ the popular NSGA-II algorithm \cite{deb2002nsga} to solve the MO problem with cheap objective functions noting that any other algorithm can be used to similar effect. NSGA-II evaluates the cheap objective functions at several inputs and sorts them into a hierarchy of sub-groups based on the ordering of Pareto dominance. The similarity between members of each sub-group and their Pareto dominance is used by the algorithm to move towards more promising parts of the input space.

\vspace{1.0ex}

\noindent {\bf Picking the best candidate input.} We need to select the best input from the Pareto set $\mathcal{X}_p$ obtained by solving the cheap MO problem. All inputs in $\mathcal{X}_p$ are promising in the sense that they represent the trade-offs in the utility space corresponding to different unknown functions. It is critical to select the input that will guide the overall search towards the goal of quickly approximating the true Pareto set $\mathcal{X}^*$. We employ a uncertainty measure defined in terms of the statistical models $\mathcal{M}_1, \mathcal{M}_2,\cdots, \mathcal{M}_k$ to select the most promising candidate input for evaluation. In single-objective optimization case, the learned model's uncertainty for an input can be defined in terms of the variance of the statistical model. For multi-objective optimization case, we define the uncertainty measure as the volume of the uncertainty hyper-rectangle.
\begin{align}
\label{eq1}
U_{\beta_t}(x) =& VOL(  \{(LCB(\mathcal{M}_i, x), UCB(\mathcal{M}_i, x) \}_{i=1}^{k} )
\end{align}

where LCB$(\mathcal{M}_i, x)$ and UCB$(\mathcal{M}_i, x)$ represent the lower confidence bound and upper confidence bound of the statistical model $\mathcal{M}_i$ for an input $x$ as defined in equations ~\ref{ucbeq} and \ref{lcbeq}; and $\beta_t$ is the parameter value to trade-off exploitation and exploration at iteration $t$. We employ the adaptive rate recommended by \cite{gp-ucb} to set the $\beta_t$ value depending on the iteration number $t$. We measure the uncertainty volume measure for all inputs $x \in \mathcal{X}_p$ and select the input with maximum uncertainty for function evaluation.
\begin{align}
    x_{t+1}=\arg max_{x\in \mathcal{X}_p} \; U_{\beta_t}(x)
\end{align}

\subsection{Theoretical Analysis}
In this section, we provide a theoretical analysis for the behavior of USeMO approach. MOO literature has multiple metrics to assess the quality of Pareto front approximation. Most commonly employed metrics include Pareto Hypervolume (PHV) indicator \cite{zitzler1999evolutionary}, $R_2$ indicator, and epsilon indicator \cite{picheny2015multiobjective}.
Both epsilon and $R_2$ metrics are instances of distance-based regret, a natural generalization of the regret measure for single-objective problems. We consider the case of LCB acquisition function and extend the cumulative regret measure for single-objective BO proposed in the well-known work by Srinivasan et al., \cite{gp-ucb} to prove convergence results. 
However, our experimental results show the generality of USeMO with different acquisition functions including TS and EI. Prior work \cite{picheny2015multiobjective} has shown that $R_2$, $epsilon$, and PHV indicator show similar behavior. Indeed, our experiments validate this claim for UseMO. We present the theoretical analysis of USeMO in terms of asymptotic regret bounds. Since the point selected in the proof is arbitrary, it holds for all points. Hence, the regret bound can be easily adapted for both epsilon and $R_2$ metrics.

Let $x^*$ be a point in the optimal Pareto set $\mathcal{X}^*$. Let $x_t$ be a point in the Pareto set $\mathcal{X}_t$ estimated by USeMO approach by solving cheap MO problem at the $t^{th}$ iteration. Let $R(x^*)$ = $\|R_1,\cdots, R_k\|$, where $R_i$ = $\sum_{t=1}^{T_{max}} (F_i(x_t)-F_i( x^*))$ 
 and $\|.\|$ is the norm of the $k$-vector and $T_{max}$ is the maximum number of iterations. We discuss asymptotic bounds for this measure using GP-LCB as an acquisition function over the input set $\mathcal{X}$. We provide proof details in \textbf{Appendix 1}.

\vspace{0.8ex}

\noindent {\bf Lemma 1} Given $\delta \in (0, 1)$ and $\beta_t$ = $2log(|\mathcal{X}|\pi^2 t^2 /6\delta)$, the following holds with probability $1-\delta$:
\begin{align}
    |F_i(x) - \mu_{i,t-1}( x)| \leq \beta_t^{1/2} \sigma_{i,t-1}(x)   \\
    \text{for all} 1 \leq i \leq k,  x \in \mathcal{X}, \text{and } t \geq 1
\end{align}

\noindent {\bf Theorem 1} If $\mathcal{X}_t$ is the Pareto set obtained by solving the cheap multi-objective optimization problem at $t$-th iteration, then the following holds with probability $1-\delta$,
\begin{align}
    R(x^*) \leq \sqrt{\sum_{i=1}^{k} {CT_{max}\beta_{T_{max}}\gamma_{T_{max}}^i }}
\end{align}\\
where $C$ is a constant and $\gamma_{T_{max}}^i$ is the maximum information gain about function $F_i$ after $T_{max}$ iterations. Essentially, this theorem suggests that since each term $R_i$ in $R(x^*)$ grows sub-linearly in the asymptotic sense, $R(x^*)$ which is defined as the norm also grows sub-linearly. 
To the best of our knowledge, this is the first work to prove a {\em sub-linear regret} for multi-objective BO setting. We proved this result using the same AF for all objectives. This is a strong theoretical-proof that USeMO is already the best in this setting. This is one of the strong reasons that justifies the use of single AF within USeMO framework.

\section{Experiments and Results}

In this section, we describe our experimental setup and present results of USeMO on diverse benchmarks.

\begin{figure*}[h!] 
    \centering
    \begin{minipage}{0.5\textwidth}
    \centering
    \begin{minipage}{0.49\textwidth}
        \centering
        \includegraphics[width=0.95\textwidth]{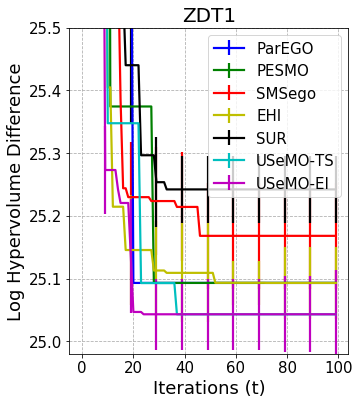} 
    \end{minipage}\hfill
    \begin{minipage}{0.49\textwidth}
        \centering
        \includegraphics[width=0.95\textwidth]{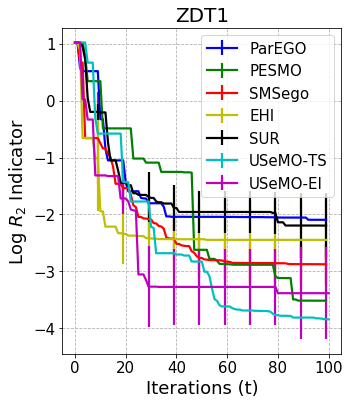} 
    \end{minipage}
    \end{minipage}\hfill
    \begin{minipage}{0.5\textwidth}
            \centering
    \begin{minipage}{0.49\textwidth}
        \centering
        \includegraphics[width=0.91\textwidth]{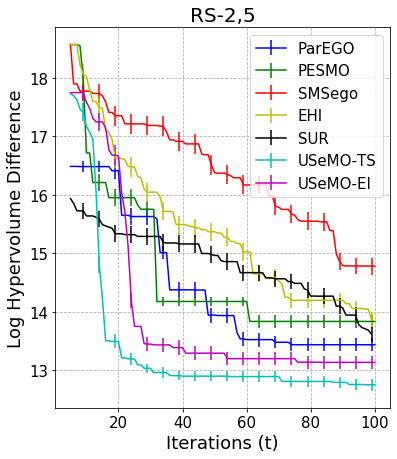} 
    \end{minipage}\hfill
    \begin{minipage}{0.49\textwidth}
        \centering
        \includegraphics[width=0.91\textwidth]{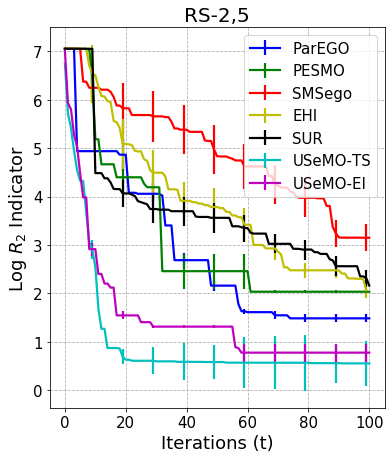} 
    \end{minipage}
    \end{minipage}
           \begin{minipage}{0.5\textwidth}
            \centering
    \begin{minipage}{0.49\textwidth}
        \centering
        \includegraphics[width=0.91\textwidth]{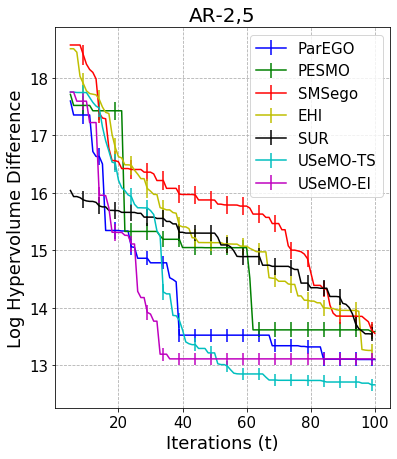} 
    \end{minipage}\hfill
    \begin{minipage}{0.49\textwidth}
        \centering
        \includegraphics[width=0.89\textwidth]{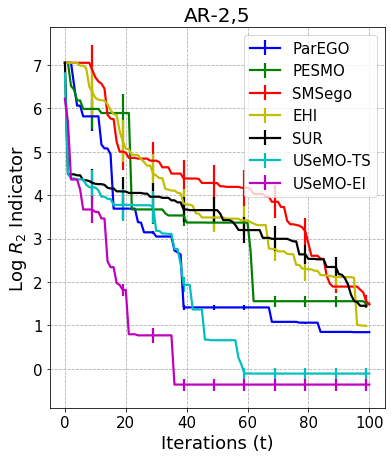} 
    \end{minipage}
    \end{minipage}\hfill
       \begin{minipage}{0.5\textwidth}
    \centering
    \begin{minipage}{0.49\textwidth}
        \centering
        \includegraphics[width=0.91\textwidth]{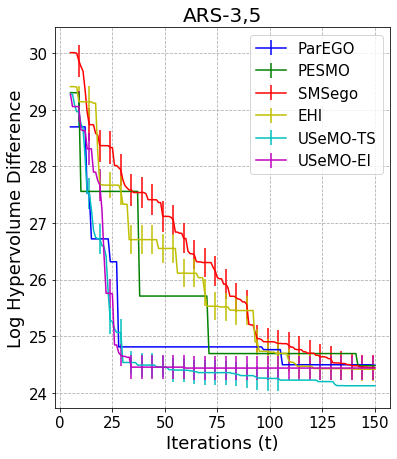} 
    \end{minipage}\hfill
    \begin{minipage}{0.49\textwidth}
        \centering
        \includegraphics[width=0.89\textwidth]{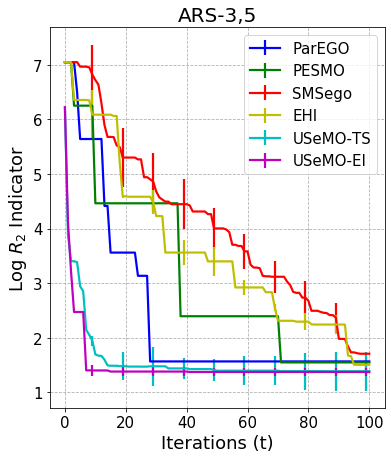} 
    \end{minipage}
    \end{minipage}
    \begin{minipage}{0.5\textwidth}
            \centering
    \begin{minipage}{0.49\textwidth}
        \centering
        \includegraphics[width=0.93\textwidth]{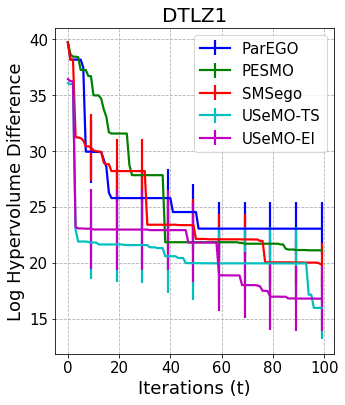} 
    \end{minipage}\hfill
    \begin{minipage}{0.49\textwidth}
        \centering
        \includegraphics[width=0.94\textwidth]{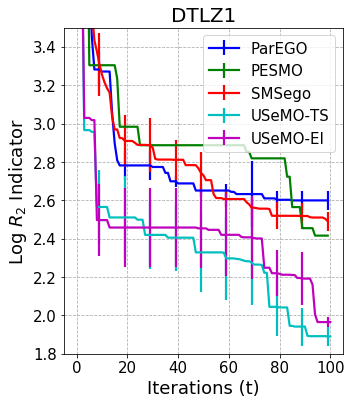} 
    \end{minipage}
    \end{minipage} \hfill
    \begin{minipage}{0.48\textwidth}
            \centering
    \begin{minipage}{0.48\textwidth}
        \centering
        \includegraphics[width=0.96\textwidth]{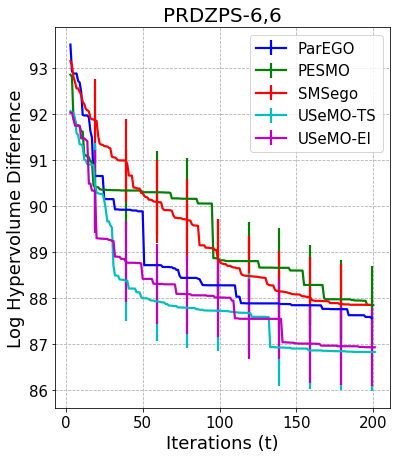} 
    \end{minipage}\hfill
    \begin{minipage}{0.48\textwidth}
        \centering
        \includegraphics[width=1\textwidth]{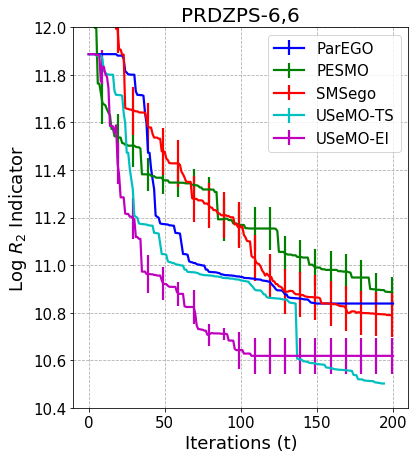} 
    \end{minipage}
    \end{minipage}

 \begin{minipage}{0.5\textwidth}
    \centering
    \begin{minipage}{0.48\textwidth}
        \centering
        \includegraphics[width=0.95\textwidth]{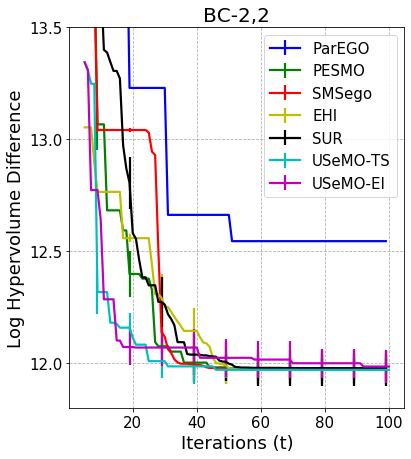} 
    \end{minipage}\hfill
    \begin{minipage}{0.48\textwidth}
        \centering
        \includegraphics[width=0.93\textwidth]{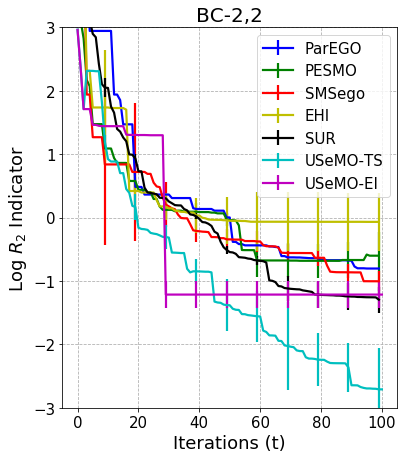} 
    \end{minipage}
    \end{minipage}\hfill
    \begin{minipage}{0.5\textwidth}
    \centering
    \begin{minipage}{0.49\textwidth}
        \centering
        \includegraphics[width=0.92\textwidth]{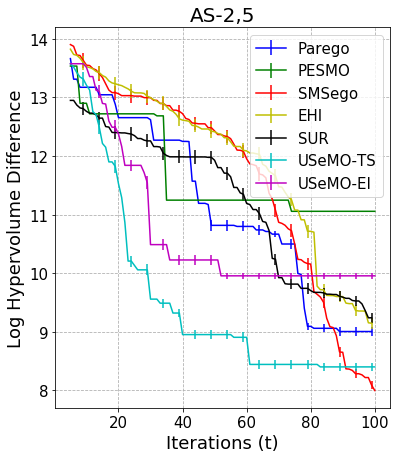} 
    \end{minipage}\hfill
    \begin{minipage}{0.48\textwidth}
        \centering
        \includegraphics[width=0.95\textwidth]{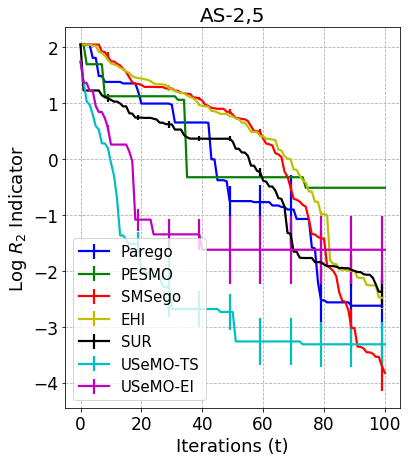} 
    \end{minipage}
    \end{minipage}
 
\caption{Results of different multi-objective BO algorithms including USeMO on synthetic benchmarks. The log of the hypervolume difference and log of $R_2$ Indicator are shown with different number of function evaluations (iterations). The mean and variance of 10 different runs are plotted. The tile of each figure refers to the benchmark name defined in Table \ref{tab:synth}.}
\label{syntheticexp}
\end{figure*}
\subsection{Experimental Setup}

\vspace{0.5ex}

\noindent {\bf Multi-objective BO algorithms.} We compare USeMO with existing methods including ParEGO \cite{knowles2006parego}, PESMO \cite{hernandez2016predictive}, SMSego \cite{ponweiser2008multiobjective}, EHI \cite{emmerich2008computation}, and SUR \cite{picheny2015multiobjective}. We employ the code for these methods from the BO library Spearmint\footnote{https://github.com/HIPS/Spearmint/tree/PESM}. We present the results of USeMO with EI and TS acquisition functions --- USeMO-TS and USeMO-EI --- noting that results show similar trend with other acquisition functions. We did not include PAL \cite{zuluaga2013active} as it is known to have similar performance as SMSego \cite{hernandez2016predictive} and works only for finite discrete input space. The code for our method is available at (github.com/belakaria/USeMO).

\vspace{0.5ex}

\noindent {\bf Statistical models.} We use a GP based statistical model with squared exponential (SE) kernel in all our experiments. 
The hyper-parameters are estimated after every 10 function evaluations. We initialize the GP models for all functions by sampling initial points at random from a Sobol grid using the in-built procedure in the Spearmint library. GPs are fitted using normalized objective function values to guarantee that all objectives are within the same range.

\vspace{0.5ex}

\noindent  {\bf Cheap MO solver.} We employ the popular NSGA-II algorithm to solve the cheap MO problem noting that other solvers can be used to similar effect.
For NSGA-II, the most important parameter is the number of function calls. We experimented with values varying from 1,000 to 20,000. We noticed that increasing this number does not result in any performance improvement for USeMO. Therefore, we fixed it to 1500 for all our experiments.

\begin{table}[h]
\centering
\resizebox{0.85\linewidth}{!}{
\begin{tabular}{llrr}  
\toprule
Name & Benchmark functions & $k$ & $d$  \\
\midrule
BC-2,2 & Branin-Currin    & 2 & 2        \\
\midrule
ZDT1 & Zitzler,Deb,Thiele  & 2  &  4     \\
\midrule
AS-2,5 & Ackley-Sphere  & 2 & 5        \\
\midrule
AR-2,5 & Ackley-Rosenbrock & 2 & 5        \\
\midrule
RS-2,5 & Rosenbrock-Sphere & 2  &  5     \\
\midrule
ARS-3,5 & Ackley-Rosenbrock-Sphere & 3 & 5        \\
\midrule
DTLZ1 & Deb,Thiele,Laumanns,Zitzler & 4 & 3        \\
\midrule
PRDZPS-6,6 & Powell-Rastrigin-Dixon & 6 & 6 \\
& Zakharov-Perm-SumSquares & &  \\
\bottomrule
\end{tabular}}
\caption{Details of synthetic benchmarks: Name, benchmark functions, no. of objectives $k$, and input dimension $d$.}
\label{tab:synth}
\end{table} 

\vspace{0.8ex}

\noindent {\bf Synthetic benchmarks.} We construct several synthetic multi-objective (MO) benchmark problems using a combination of commonly employed benchmark function for single-objective optimization\footnote{https://www.sfu.ca/~ssurjano/optimization.html} and two of the known general MO benchmarks. We provide the complete details of these MO benchmarks in Table~\ref{tab:synth}. Due to space constraints we present some of the results in the appendix 
\begin{figure*}[h!] 
    \centering
    \begin{minipage}{0.5\textwidth}
    \centering
    \begin{minipage}{0.5\textwidth}
        \centering
        \includegraphics[width=0.95\textwidth]{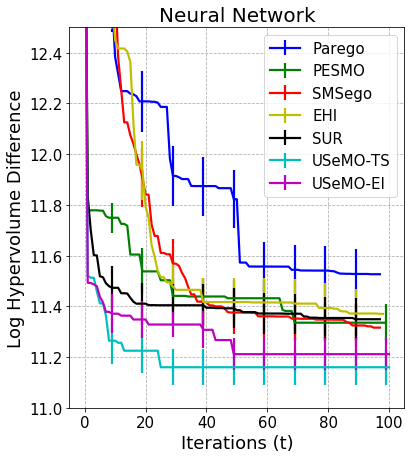} 
    \end{minipage}\hfill
    \begin{minipage}{0.5\textwidth}
        \centering
        \includegraphics[width=0.95\textwidth]{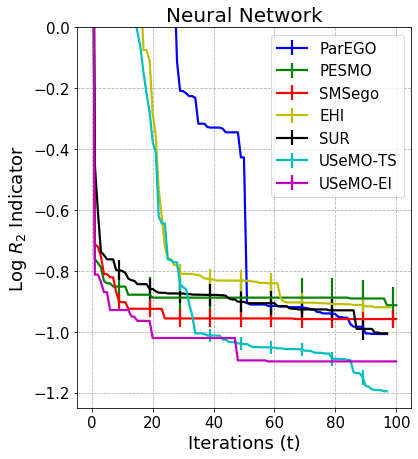} 
    \end{minipage}
    \end{minipage}\hfill
    \begin{minipage}{0.5\textwidth}
            \centering
    \begin{minipage}{0.5\textwidth}
        \centering
        \includegraphics[width=0.91\textwidth]{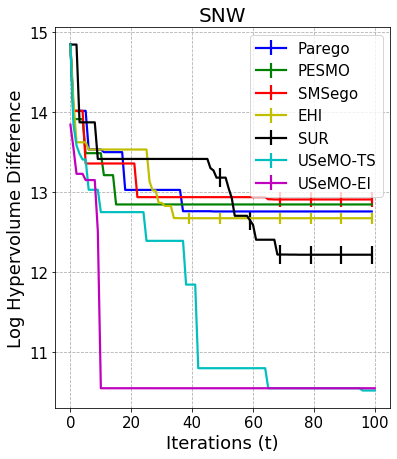} 
    \end{minipage}\hfill
    \begin{minipage}{0.5\textwidth}
        \centering
        \includegraphics[width=0.91\textwidth]{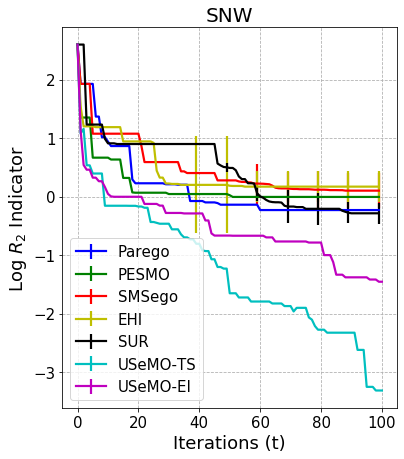} 
    \end{minipage}
    \end{minipage}
    \begin{minipage}{0.5\textwidth}
    \centering
    \begin{minipage}{0.5\textwidth}
        \centering
        \includegraphics[width=0.95\textwidth]{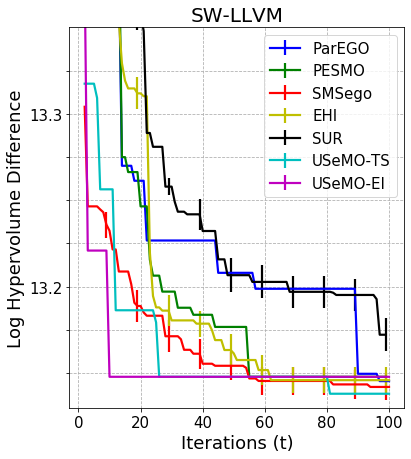} 
    \end{minipage}\hfill
    \begin{minipage}{0.5\textwidth}
        \centering
        \includegraphics[width=0.95\textwidth]{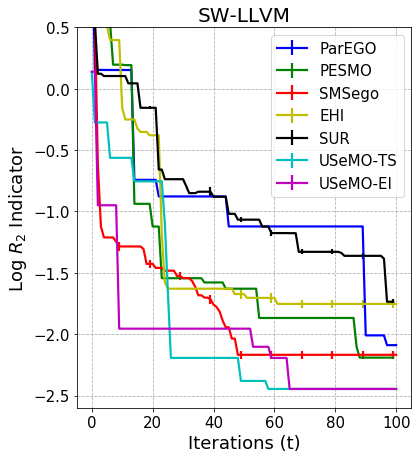} 
    \end{minipage}
    \end{minipage}\hfill
    \begin{minipage}{0.5\textwidth}
            \centering
    \begin{minipage}{0.5\textwidth}
        \centering
        \includegraphics[width=0.91\textwidth]{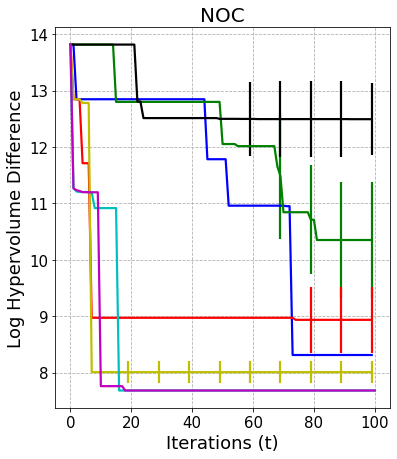} 
    \end{minipage}\hfill
    \begin{minipage}{0.5\textwidth}
        \centering
        \includegraphics[width=0.91\textwidth]{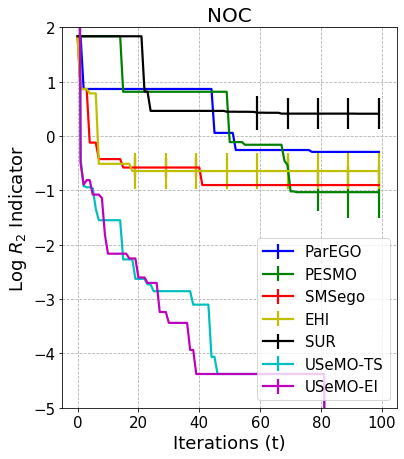} 
    \end{minipage}
    \end{minipage}
    
    \begin{minipage}{0.5\textwidth}
    \centering
    \begin{minipage}{0.5\textwidth}
        \centering
        \includegraphics[width=0.92\textwidth]{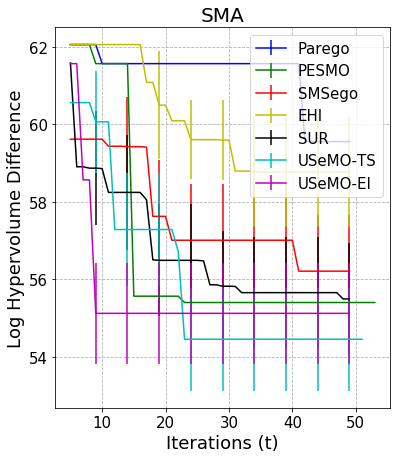} 
    \end{minipage}\hfill
    \begin{minipage}{0.5\textwidth}
        \centering
        \includegraphics[width=0.92\textwidth]{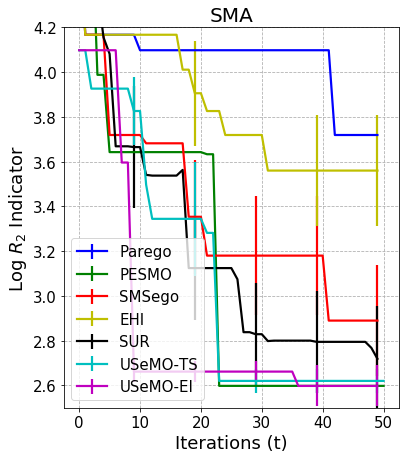} 
    \end{minipage}
    \end{minipage}\hfill
    \begin{minipage}{0.5\textwidth}
            \centering
    \begin{minipage}{0.5\textwidth}
        \centering
        \includegraphics[width=0.9\textwidth]{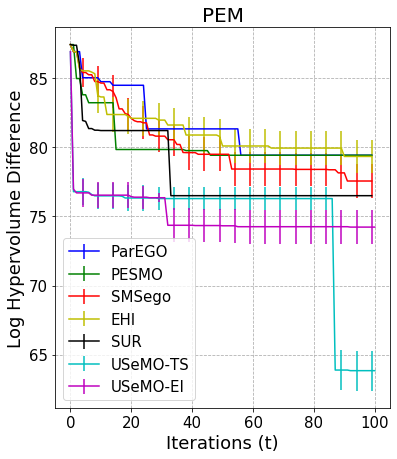} 
    \end{minipage}\hfill
    \begin{minipage}{0.5\textwidth}
        \centering
        \includegraphics[width=0.9\textwidth]{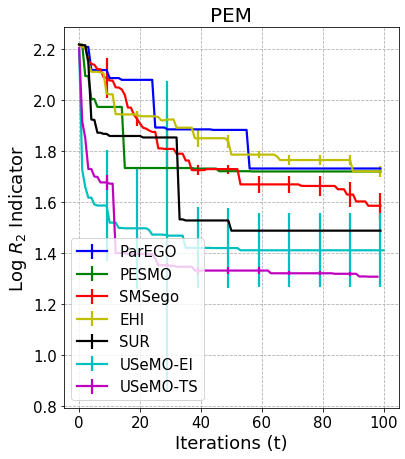} 
    \end{minipage}
    \end{minipage}

\caption{Results of different multi-objective BO algorithms including USeMO on real-world benchmarks. The log of the hypervolume difference and Log $R_2$ Indicator are shown with different number of function evaluations (iterations). The mean and variance of 10 different runs are plotted. The tile of each figure refers to the name of real-world benchmarks.}
\label{realworldexp}
\end{figure*}
\vspace{0.8ex}

\noindent {\bf Real-world benchmarks.} We employed six diverse real-world benchmarks for our experiments. 

{\bf 1) Hyper-parameter tuning of neural networks.} 
Our goal is to find a neural network with high accuracy and low prediction time. 
We optimize a dense neural network over the MNIST dataset \cite{lecun1998gradient}.
Hyper-parameters include the number of hidden layers, the number of neurons per layer, the dropout probability, the learning rate, and the regularization weight penalties $l_1$ and $l_2$. We employ 10K instances for validation and 50K instances for training. We train the network for 100 epochs for evaluating each candidate hyper-parameter values on validation set. We apply a logarithm function to error rates due to their small values.

{\bf 2) SW-LLVM compiler settings optimization.} SW-LLVM is a data set with 1024 compiler settings \cite{siegmund2012predicting} determined by $d$=10 binary inputs. The goal of this experiment is to find a setting of the LLVM compiler that optimizes the memory footprint and performance on a given set of software programs. Evaluating these objectives is very costly and testing all the settings takes over 20 days.

{\bf 3) SNW sorting network optimization.} The data set SNW was first introduced by \cite{zuluaga2012computer}. The goal is to optimize the area and throughput for the synthesis of a field-programmable gate array (FPGA) platform. The input space consists of 206 different hardware design implementations of a sorting network. Each design is defined by $d$ = 4 input variables.

{\bf 4) Network-on-chip (NOC) optimization.} The design space of NoC dataset \cite{almer2011learning} consists of 259 implementations of a tree-based network-on-chip. Each configuration is defined by $d$ = 4 variables: width, complexity, FIFO, and multiplier. We optimize energy and runtime of application-specific integrated circuits (ASICs)
on the Coremark benchmark workload.

{\bf 5) Shape memory alloys (SMA) optimization.} The materials dataset SMA consists of 77 different design configurations of shape memory alloys \cite{gopakumar2018multi}. The goal is to optimize thermal hysteresis and transition temperature of alloys. Each design is defined by $d$ = 6 input variables (e.g., atomic size of the alloying elements including metallic radius and valence electron number).

{\bf 6) Piezo-electric materials (PEM) optimization.} PEM is a materials dataset consisting of 704 configurations of Piezoelectric materials \cite{gopakumar2018multi}. The goal is to optimize piezoelectric modulus and bandgap of these material designs. Each design configuration is  defined by $d$ = 7 input variables (e.g., ionic radii, volume, and density).

\vspace{0.8ex}

\noindent {\bf Evaluation metrics.}  We employ two common metrics. The {\em Pareto hypervolume (PHV)} metric is commonly employed to measure the quality of a given Pareto front \cite{zitzler1999evolutionary}. PHV is defined as the volume between a reference point and the given Pareto front (set of non-dominated points). After each iteration $t$ 
, we report the difference between the hypervolume of the ideal Pareto front $(\mathcal{Y^*}) $ and hypervolume of the estimated Pareto front $(\mathcal{Y}_{t})$ by a given algorithm. 
The {\em $R_2$ Indicator} is the average distance between the ideal Pareto front $(\mathcal{Y^*}) $ and the estimated Pareto front $(\mathcal{Y}_{t})$ by a given algorithm \cite{picheny2015multiobjective}. The $R_2$ metric degenerates to the regret metric presented in our theoretical analysis. 

\subsection{Results and Discussion}
\vspace{0.2ex}
{\bf USeMO vs. State-of-the-art.} We evaluate the performance of USeMO with different acquisition functions including TS, EI, and LCB. Due to space constraints, we show the results for USeMO with TS and EI, two very different acquisition functions, to show the generality and robustness of our approach. We also provide more results with LCB acquisition function in Appendix. Figure \ref{syntheticexp} and Figure \ref{realworldexp} show the results of all multi-objective BO algorithms including USeMO for synthetic and real-world benchmarks respectively. We make the following empirical observations: 1) USeMO consistently performs better than all baselines and also converges much faster. For blackbox optimization problems with expensive function evaluations, faster convergence has practical benefits as it allows the end-user or decision-maker to stop early. 2) Rate of convergence of USeMO varies with different acquisition functions (i.e., TS and EI), but both cases perform better than baseline methods. 3) The convergence rate of PESMO becomes slower as the dimensionality of input space grows for a fixed number of objectives, whereas USeMO maintains a consistent convergence behavior. 4) Performance of ParEGO is very inconsistent. In some cases, it is comparable to USeMO, but performs poorly on many other cases. This is expected due to random scalarization.

\vspace{0.5ex}

\begin{figure} [h!]
\centering
\begin{minipage}{.24\textwidth}
  \centering
  \includegraphics[width=1\linewidth]{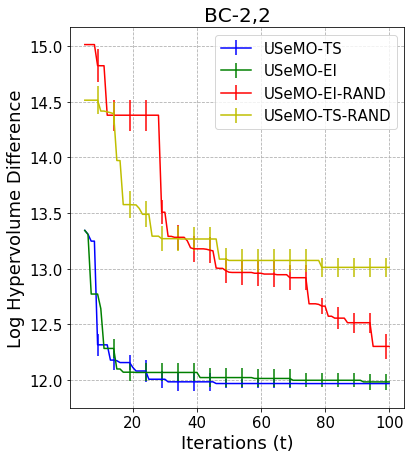}
\end{minipage}%
\begin{minipage}{.23\textwidth}
  \centering
  \includegraphics[width=1\linewidth]{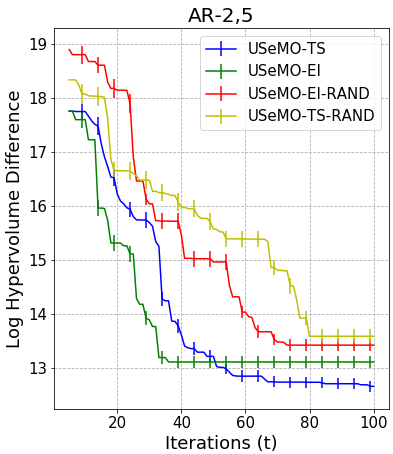}
\end{minipage}
\begin{minipage}{.23\textwidth}
  \centering
  \includegraphics[width=1\linewidth]{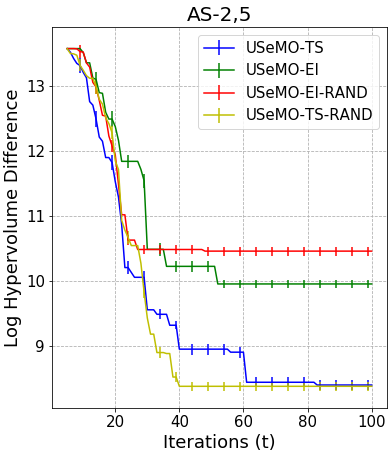}
\end{minipage}%
\begin{minipage}{.235\textwidth}
  \centering
  \includegraphics[width=1\linewidth]{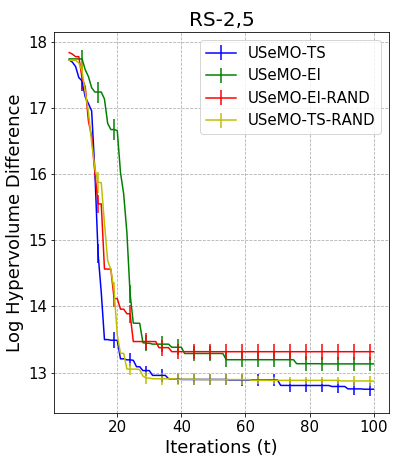}
\end{minipage}\hfill
\begin{minipage}{.23\textwidth}
  \centering
  \includegraphics[width=1\linewidth]{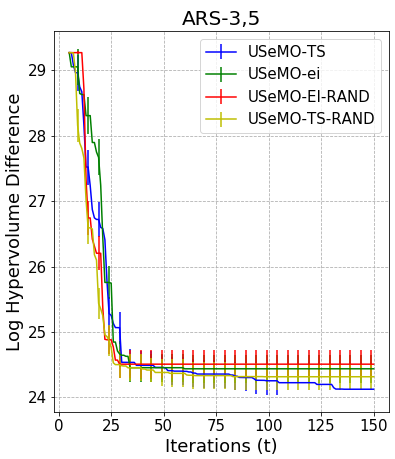}
\end{minipage}%
\begin{minipage}{.235\textwidth}
  \centering
  \includegraphics[width=1.\linewidth]{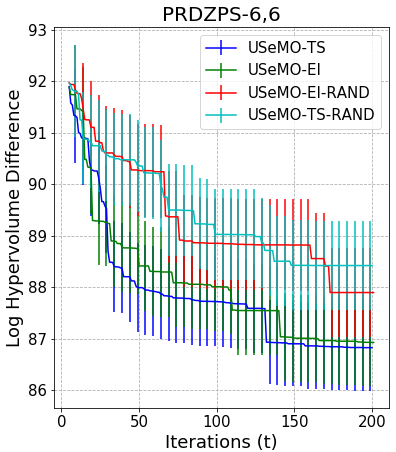}
\end{minipage}
\caption{Comparison of USeMO with uncertainty maximization and random policy for selecting the best input from Pareto set obtained by solving cheap MO problem. We plot the log of the hypervolume difference for several synthetic benchmark problems as a function of the number of evaluations. The mean and variance of 10 different runs are plotted. The figure title refers to the benchmark name defined in table \ref{tab:synth}. (Better seen in color).}\label{random}
\end{figure}

\noindent {\bf Uncertainty maximization vs. random selection.} Recall that USeMO needs to select one input for evaluation from the promising candidates obtained by solving a cheap MO problem. We compare uncertainty maximization and random policy for selection in figure \ref{random} . We observe that uncertainty maximization performs better than random policy. However, in some cases, random policy is competitive, which shows that all candidates from the solution of cheap MO problem are promising and improve the efficiency.  

\vspace{0.5ex}

\noindent {\bf Comparison of acquisition function optimization time.} We compare the runtime of acquisition function optimization for different multi-objective BO algorithms including USeMO. We do not account for the time to fit GP models since it is same for all the algorithms. We measure the average acquisition function optimization time across all iterations. we run all experiments on a machine with the following configuration: Intel i7-7700K CPU @ 4.20GHz with 8 cores and 32 GB memory. Table~\ref{tablerun} shows the time in seconds for synthetic benchmarks. We can see that USeMO scales significantly better than state-of-the-art method PESMO. USeMO is comparable to ParEGO, which relies on scalarization to reduce to acquisition optimization in single-objective BO. The time for PESMO and SMSego increases significantly as the number of objectives grow beyond two. 
\begin{table}[h]
\centering
\resizebox{1\linewidth}{!}{
\begin{tabular}{l|llllll}  
\toprule
\diaghead{\theadfont Diag ColumnmnHead II}
{Benchmarks}{MO Algorithms} & USeMO & PESMO & ParEGO & SMSego\\
\midrule
BC-2,2 & $ 4.1\pm0.7 $ & 13.6$\pm$3.2 &  4.2$\pm$ 1.6 & 80.5$\pm$ 2.1\\
\midrule
ZDT1 & $ 5\pm0.3 $ & 14.1$\pm$2.1 &  4.8$\pm$ 1.2 & 84$\pm$ 6.7\\
\midrule
RS-2,5 & 5.3$\pm$1.4 & 16.9$\pm$1.9 & 5.7$\pm$ 1.1&   90.2$\pm$8.2 \\
\midrule
ARS-3,5& 7.0 $\pm$1.5 &  34.8$\pm$12.6 & 6.7$\pm$1.4 &  135.0 $\pm$12.4\\
\midrule
DTLZ1 & 9 $\pm$2.4 &  63.6$\pm$10.1 & 8.2$\pm$0.9 &  215 $\pm$16.2\\
\midrule
PRDZPS-6,6 & 13.9$\pm$ 1.1  & 110.4$\pm$17.8 & 12.3 $\pm$ 2.3& 300.43 $\pm$ 35.7 \\
\bottomrule
\end{tabular}
}
\caption{Acquisition function optimization time in secs.}
\label{tablerun}
\end{table}

\section{Summary and Future Work}

We introduced a novel 
framework referred as USeMO to solve multi-objective Bayesian optimization problems. The key idea is a two-stage search procedure to improve the accuracy and efficiency of sequential decision-making under uncertainty for selecting inputs for evaluation. Our experimental results on
diverse benchmarks showed that USeMO yields consistently better results than state-of-the-art methods and scales gracefully to large-scale MO problems. Future work includes using USeMO to solve novel engineering and scientific applications \cite{DATE-2020}.

\vspace{1.0ex}

\noindent {\bf Acknowledgements.} This research is supported by National Science Foundation grants IIS-1845922 and OAC-1910213.

\input{main.bbl}
\clearpage
\appendix
\input{appendix.tex}
\end{document}

%% file: appendix.tex
\section{Theoretical Analysis}
This section provides the proof of Theorem 1 which depends on Lemma 1.
For completeness, the GP-LCB acquisition function uses the following definition of $LCB_{i,t}(x)$ for function $F_i$ at any iteration $t$.
\begin{align}
    LCB_{i,t}(x) = \mu_{i,t-1}(x) - \beta^{1/2}_t \sigma_{ i,t-1}( x)
\end{align}

\noindent {\bf Lemma 1} Given $\delta \in (0, 1)$ and $\beta_t = 2 \log(|\mathcal{X}|\pi^2 t^2 /6\delta)$, the following holds with probability $1-\delta$:
\begin{align}
    |F_i(x) - \mu_{i,t-1}( x)| \leq \beta_t^{1/2} \sigma_{i,t-1}(x)   \\
    \text{for all } 1 \leq i \leq k,  x \in \mathcal{X}, \text{for all } t \geq 1
\end{align}

\noindent  {\bf Proof}. Since $F_i$ is modeled by a GP, $F_i(x) \sim \mathcal{N}(\mu_{i,t-1}(x), \sigma_{i,t-1}(x))$. According to Lemma 5.1 from \cite{gp-ucb}, the below inequality holds:
\begin{align}
    \Pr \{ |F_i(x) - \mu_{i,t-1}(x)| > \beta_t^{1/2}\sigma_{i,t-1}(x) \} \leq e^{-\beta/2}
\end{align}
Applying the union bound,
\begin{align}
        |F_i(x) - \mu_{i,t-1}( x)| \leq \beta_t^{1/2} \sigma_{i,t-1}(x)
\end{align}
holds with probability $1-\delta$.
The lemma holds by choosing $e^{-\beta/2}  |\mathcal{X}| = 6\, \delta /\pi^2 t^2$ as suggested in \cite{gp-ucb}.

\vspace{1ex}
\noindent {\bf Theorem 1} If $\mathcal{X}_{t}$ be the Pareto-set generated by the cheap multi-objective optimization at $t$-th iteration, then the following holds with probability $1-\delta$,
\begin{align}
    R(x^*) \leq \sqrt{\sum_{i=1}^k {CT_{max}\beta_{T_{max}}\gamma_{T_{max}}^i }}
\end{align}\\
where $C$ is a constant and $\gamma_{T_{max}}^i$ is the maximum information gain about $F_i$ after ${T_{max}}$ iterations.
\vspace{1ex}

\noindent {\bf Proof.} \noindent For the sake of completeness, the cheap multi-objective optimization problem for GP-LCB becomes
\begin{align}
     \min_{x \in \mathcal{X}}\, (LCB_{1,t}(x), LCB_{2,t}(x),\cdots, LCB_{k,t}(x))
\end{align}
Assuming optimality of $\mathcal{X}_{t}$, either  there exists a $x_t \in \mathcal{X}_{t}$ such that 
\begin{align}
LCB_{i,t} (x_t) \leq LCB_{i,t} (x^*), \forall{i \in \{1,\cdots,k\}}
\end{align}
or $x^*$ is in the optimal Pareto set  $\mathcal{X}_{t}$ generated by cheap MO solver (i.e., $x_t = x^*$).
\vspace{1ex}

\noindent Now, using Lemma 1 for any function $F_j$, 
\begin{align}
LCB_{j,t} (x_t) \leq LCB_{j,t} (x^*) \leq F_j(x^*)
\end{align}
Therefore,
\begin{align}
    R_j(x^*) &= F_j(x_t) - F_j(x^*) \leq F_j(x_t) - LCB_{j,t} (x_t) \\
    R_j(x^*) &\leq F_j(x_t) - \mu_{i,t-1}(x_t) + \beta^{1/2} \sigma_{i,t-1}(x_t) \\
     R_j(x^*) &\leq 2\beta^{1/2} \sigma_{i,t-1}(x_t)\label{equation:1}
\end{align}
Inequality (\ref{equation:1}) is similar to the result of Lemma 5.2 from \cite{gp-ucb} in the single-objective BO case. Since $j$ is arbitrary, this is true for each function $F_j$, for all $j \in \{1, 2,\cdots, k\}$. \\

\noindent Further, using Lemma 5.4 from \cite{gp-ucb},
\begin{align}
    R_j(x^*) \leq \sqrt{C T_{max} \beta_{T_{max}} \gamma_{T_{max}}^j} \text{ {with probability}} \geq 1-\delta 
\end{align}
Consequently, the bounds for $R(x^*)$ becomes
\begin{align}
    R(x^*) \leq \sqrt{\sum_{i=1}^k {CT_{max}\beta_{T_{max}}\gamma_{T_{max}}^i }}
\end{align}
The quantity $\gamma_{T_{max}}^i$ is employed in many theoretical studies of GP based optimization including the well-known work of Srinivasan et al., \cite{gp-ucb}.

\section{Additional Experimental Results}
\begin{figure}[h]
    \centering
    \includegraphics[scale=0.5]{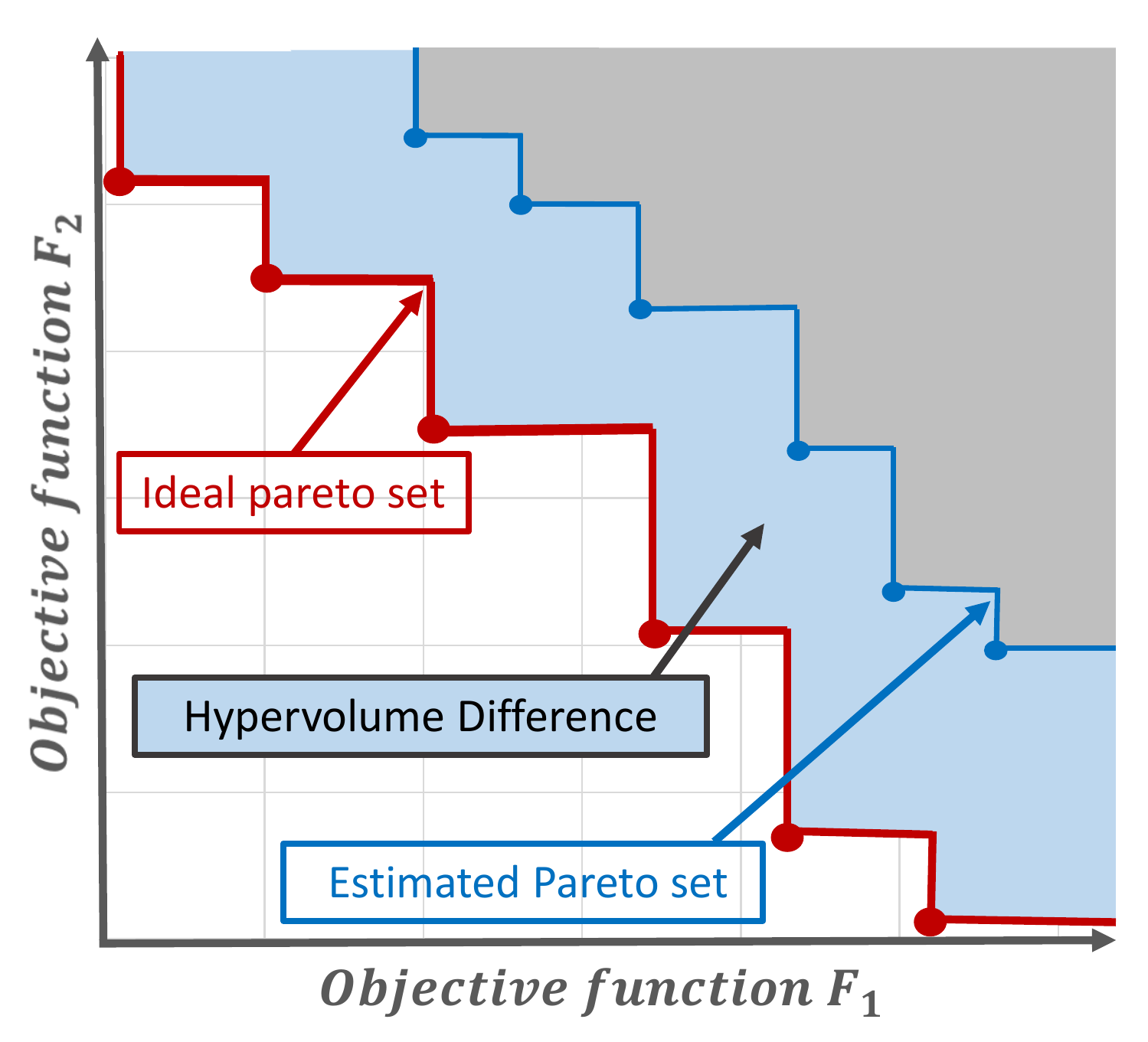}
    \caption{Illustration of Pareto hypervolume difference. The blue points correspond to the Pareto front estimated by a given algorithm. The gray volume is its corresponding Pareto hypervolume. The red points are correspond to the optimal Pareto front. The blue area represents the Pareto hypervolume difference metric for this example.}
    \label{hypervolume}
\end{figure}

Figure \ref{hypervolume} provides an illustration of the hypervolume measure.\\

\begin{figure*}[h]
\centering
\begin{minipage}{.4\textwidth}
  \centering
  \includegraphics[width=1\linewidth]{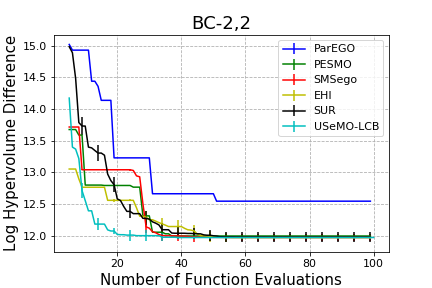}
  \label{fig:test1}
\end{minipage}%
\begin{minipage}{.4\textwidth}
  \centering
  \includegraphics[width=1\linewidth]{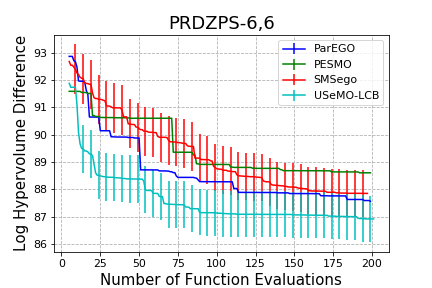}
  \label{fig:test2}
\end{minipage}
\begin{minipage}{.4\textwidth}
  \centering
  \includegraphics[width=1\linewidth]{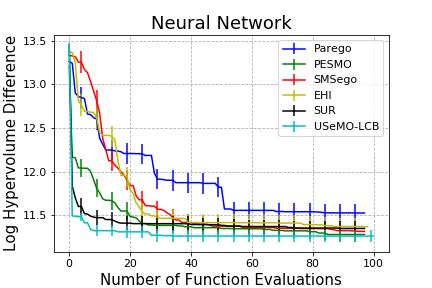}
  \label{fig:test3}
\end{minipage}%
\begin{minipage}{.4\textwidth}
  \centering
  \includegraphics[width=1\linewidth]{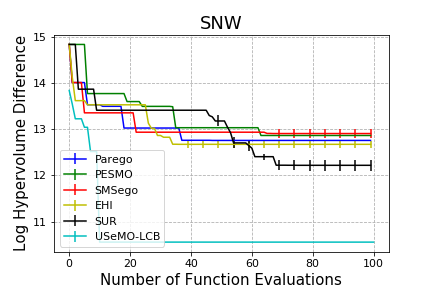}
  
\end{minipage}

\caption{Comparison of USeMO-LCB with baseline algorithms. We plot the log of the hypervolume difference for synthetic and real-world benchmark problems as a function of the number of evaluations. The mean and variance of 10 different runs are plotted. The figure title refers to the benchmark name defined in the experiments section. (Better seen in color).}\label{AppendLCB}
\end{figure*}

\noindent Figure \ref{AppendLCB} shows the results of USeMO with LCB acquisition function and comparison with baseline multi-objective BO algorithms. \\

